%% file: main.tex
\documentclass[sigconf]{acmart}
\AtBeginDocument{%
  }

\usepackage{booktabs}
\usepackage{multirow}
\usepackage{makecell}
\usepackage{xspace}
\usepackage{graphicx}
\usepackage{enumitem}
\usepackage{fontawesome}
\usepackage[most]{tcolorbox}
\usepackage{tabularx}
\usepackage{xcolor}
\usepackage{hyperref}
\newcolumntype{Y}{>{\centering\arraybackslash}X}

\newcommand{\nyx}{\textbf{\textsc{Nyx}}\xspace}
\newcommand{\nyxqa}{\textbf{\textsc{NyxQA}}\xspace}

\setcopyright{acmlicensed}
\copyrightyear{2026}
\acmYear{2026}
\acmDOI{XXXXXXX.XXXXXXX}
\acmConference[WWW '26]{ACM The Web Conference 2026}{April 13--17, 2026}{Dubai, UAE}
\acmBooktitle{Proceedings of the ACM Web Conference 2026}
\acmISBN{978-1-4503-XXXX-X/2018/06}

\begin{document}

\title{Towards Mixed-Modal Retrieval for Universal Retrieval-Augmented Generation}

\author{Chenghao Zhang\\Guanting Dong}
\affiliation{
  \institution{Renmin University of China}
  \city{Beijing}
  \country{China}
}
\email{chenghao-zhang@outlook.com}

\author{Xinyu Yang\\Zhicheng Dou}
\affiliation{
  \institution{Renmin University of China}
  \city{Beijing}
  \country{China}
}
\email{dou@ruc.edu.cn}

\renewcommand{\shortauthors}{Zhang et al.}

\begin{abstract}

Retrieval-Augmented Generation (RAG) has emerged as a powerful paradigm for enhancing large language models (LLMs) by retrieving relevant documents from an external corpus. However, existing RAG systems primarily focus on unimodal text documents, and often fall short in real-world scenarios where both queries and documents may contain mixed modalities (such as text and images). In this paper, we address the challenge of Universal Retrieval-Augmented Generation (URAG), which involves retrieving and reasoning over mixed-modal information to improve vision-language generation. To this end, we propose \nyx, a unified mixed-modal to mixed-modal retriever tailored for URAG scenarios. To mitigate the scarcity of realistic mixed-modal data, we introduce a four-stage automated pipeline for data generation and filtering, leveraging web documents to construct \nyxqa, a dataset comprising diverse mixed-modal question-answer pairs that better reflect real-world information needs. Building on this high-quality dataset, we adopt a two-stage training framework for \nyx: we first perform pre-training on \nyxqa along with a variety of open-source retrieval datasets, followed by supervised fine-tuning using feedback from downstream vision-language models (VLMs) to align retrieval outputs with generative preferences. Experimental results demonstrate that \nyx not only performs competitively on standard text-only RAG benchmarks, but also excels in the more general and realistic URAG setting, significantly improving generation quality in vision-language tasks. Our code is released at \href{https://github.com/SnowNation101/Nyx}{\textcolor{cyan}{https://github.com/SnowNation101/Nyx}}

\end{abstract}


\settopmatter{printacmref=false}



\maketitle

\section{Introduction}

\begin{figure}[t]
    \centering
    \includegraphics[width=0.5\textwidth]{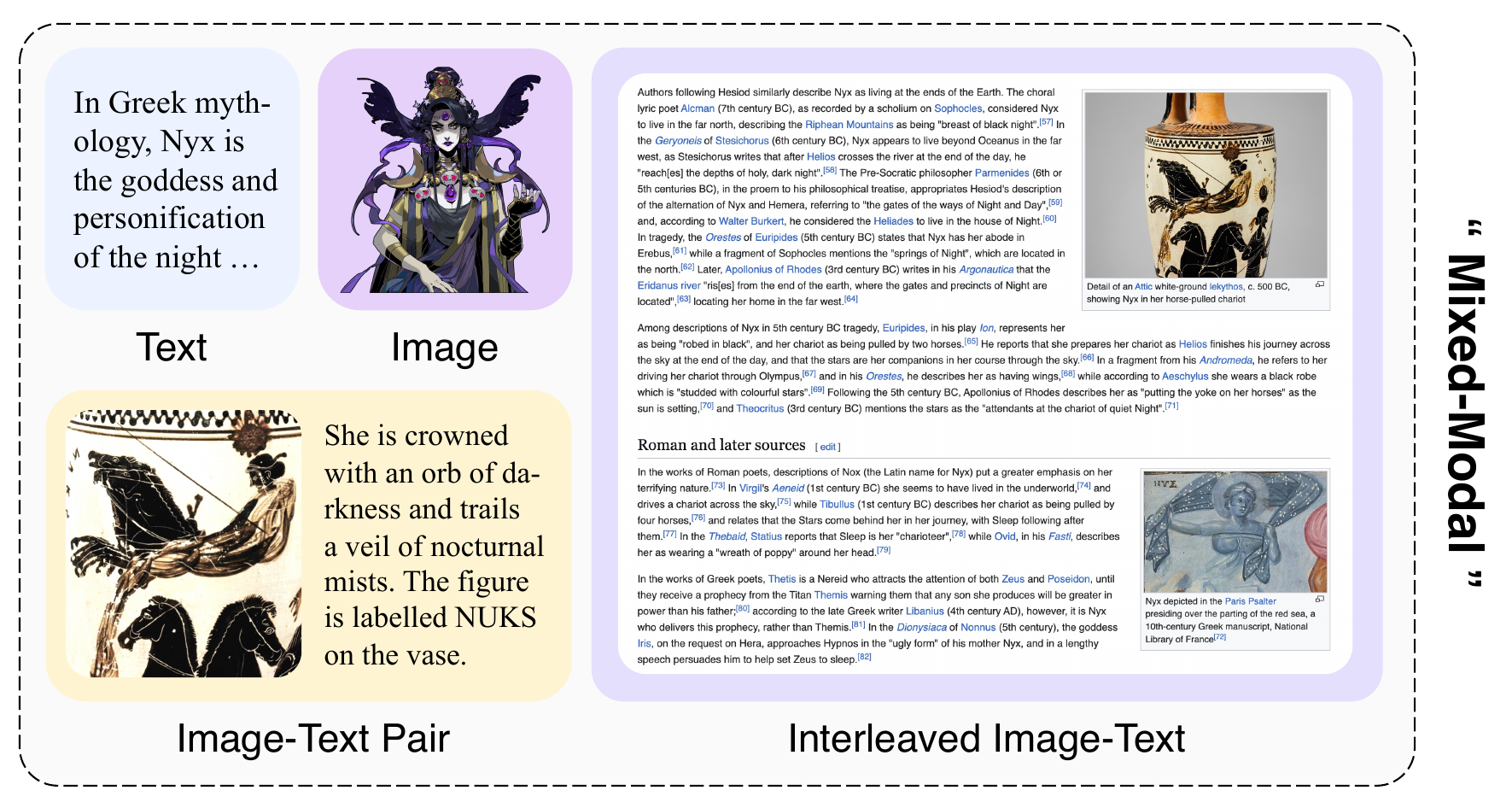}
    \caption{An illustration of the input patterns of ``mixed-modal'' content in the URAG scenario.}
    \Description{}
    \label{fig:example}
\end{figure}

Large language models (LLMs) have shown remarkable capabilities in text comprehension and generation~\cite{meta2024llama3,  tongyi2025qwen3, deepseekai2024dsv3,qiao2024wemath,qiao2025we}. To extend their capabilities to multimodal understanding, vision-language models (VLMs) incorporate visual encoders to process text and image inputs~\cite{tongyi2025qwen25vl, zhu2025internvl3}. However, like LLMs, VLMs often struggle with queries needing up-to-date or external knowledge. Retrieval-Augmented Generation (RAG) addresses this by retrieving documents from an external corpus to complement internal knowledge~\cite{lewis2025rag, dong2025dparag,dong2025tool}. Building on this, Multimodal RAG (MRAG) extends the paradigm to settings where both queries and documents may contain text, images, or both~\cite{chen2022murag, yu2025visrag}.

Current MRAG methods fall broadly into two categories: \textbf{(1) The divide-and-conquer approach}, which utilize text queries for text documents and visual queries for images; \textbf{(2) The cross-modal retrieval}, which uses text queries to retrieve visual content.  However, both paradigms suffer from notable limitations. They often overlook the spatial and logical relationships between images and text within a document, making it difficult to capture fine-grained interactions crucial for downstream reasoning.

However, web documents in the real world are often far more complex and diverse. As illustrated in Figure~\ref{fig:example}, they may include pure text, individual images, paired image-text content, or arbitrarily interleaved sequences of text and images. We refer to this broad spectrum of formats as \textit{mixed-modal} content, where the interplay between modalities plays a critical role in conveying meaning.

While recent efforts, such as VLM2Vec~\cite{jiang2025vlm2vec}, have introduced unified multimodal embedding models, these approaches mainly focus on embedding pure text, individual images, or neatly aligned text-image pairs. Consequently, they face challenges in handling complex multimodal structures, such as interwoven or densely intertwined text and images. Furthermore, their application within the MRAG framework is still largely unexplored.

\textbf{This gap becomes even more pronounced in the context of Universal Retrieval-Augmented Generation (URAG)}, as both queries and documents can be of arbitrary mixed modalities. Unlike traditional settings with purely textual queries, URAG introduces dual challenges: understanding heterogeneous inputs and retrieving from equally diverse corpora. An effective retriever needs to be capable of encoding various content types and matching them with complex document structures. This imposes new technical requirements on representation learning, matching precision, and alignment with downstream VLMs, highlighting the need for a truly universal, flexible, and vision-language-aware retrieval paradigm.

To address these challenges, we propose \nyx, a unified retriever designed for mixed-modal-to-mixed-modal retrieval in URAG scenarios. To mitigate data scarcity of realistic URAG training data, we first introduce a four-stage automatic pipeline to build \nyxqa, a new dataset tailored for URAG. \nyxqa consists of three components: (1) a large-scale mixed-modal multiple-choice question answering (QA) dataset, (2) a corresponding mixed-modal document corpus, and (3) a pretraining dataset for contrastive learning.

Our construction process begins with sampling naturally interleaved image-text documents from the web to form the corpus. We then employ a powerful VLM to generate QA pairs conditioned on these documents. To ensure high data quality, we apply a multi-step post-processing procedure, yielding a clean and diverse QA set. Finally, based on these QA pairs, we mine hard negatives from the corpus to form the pretraining triplets used for contrastive learning. Unlike existing multimodal datasets limited to specific modality combinations, \nyxqa supports retrieval and generation involving arbitrarily structured text, images, and their interleaved formats.

Building upon this dataset, we adopt a two-stage training framework to develop \nyx from a pretrained VLM. In the first stage, we pretrain the retriever on \nyxqa and several public contrastive learning datasets to establish general-purpose multimodal retrieval capabilities. To balance retrieval effectiveness and efficiency, we incorporate Matryoshka Representation Learning (MRL)~\cite{kusupati2022mrl}, resulting in a compact yet expressive encoder, termed \nyx-pretrained. In the second stage, we perform feedback-driven fine-tuning, aligning the retriever with the generative preferences of downstream VLMs. This yields the final version of our retriever, \nyx.

Extensive experiments demonstrate that \nyx consistently enhances retrieval accuracy and downstream reasoning performance in challenging mixed-modal scenarios, showcasing its strong suitability for URAG tasks. 

In conclusion, our contributions are as follows:
\begin{itemize}
\item We pioneer the exploration of the Universal Retrieval Augmented Generation (URAG) problem, addressing scenarios where both queries and documents consist of arbitrarily interleaved image-text content.

\item We introduce a dataset specifically designed for real-world URAG applications, created through a comprehensive four-step web-based multimodal data synthesis pipeline. This dataset offers a rich variety of interleaved content formats, serving as an effective benchmark for practical multimodal retrieval tasks.

\item We propose a two-stage training paradigm to develop \nyx, a unified retriever optimized for URAG. The first stage involves contrastive pretraining using MRL on both public and synthetic datasets, resulting in \nyx-pretrained. Then we utilize feedback from VLMs to refine the retriever through targeted supervision, culminating in our final model, \nyx.
\end{itemize}
\begin{figure*}[ht]
    \centering
    \includegraphics[width=\textwidth]{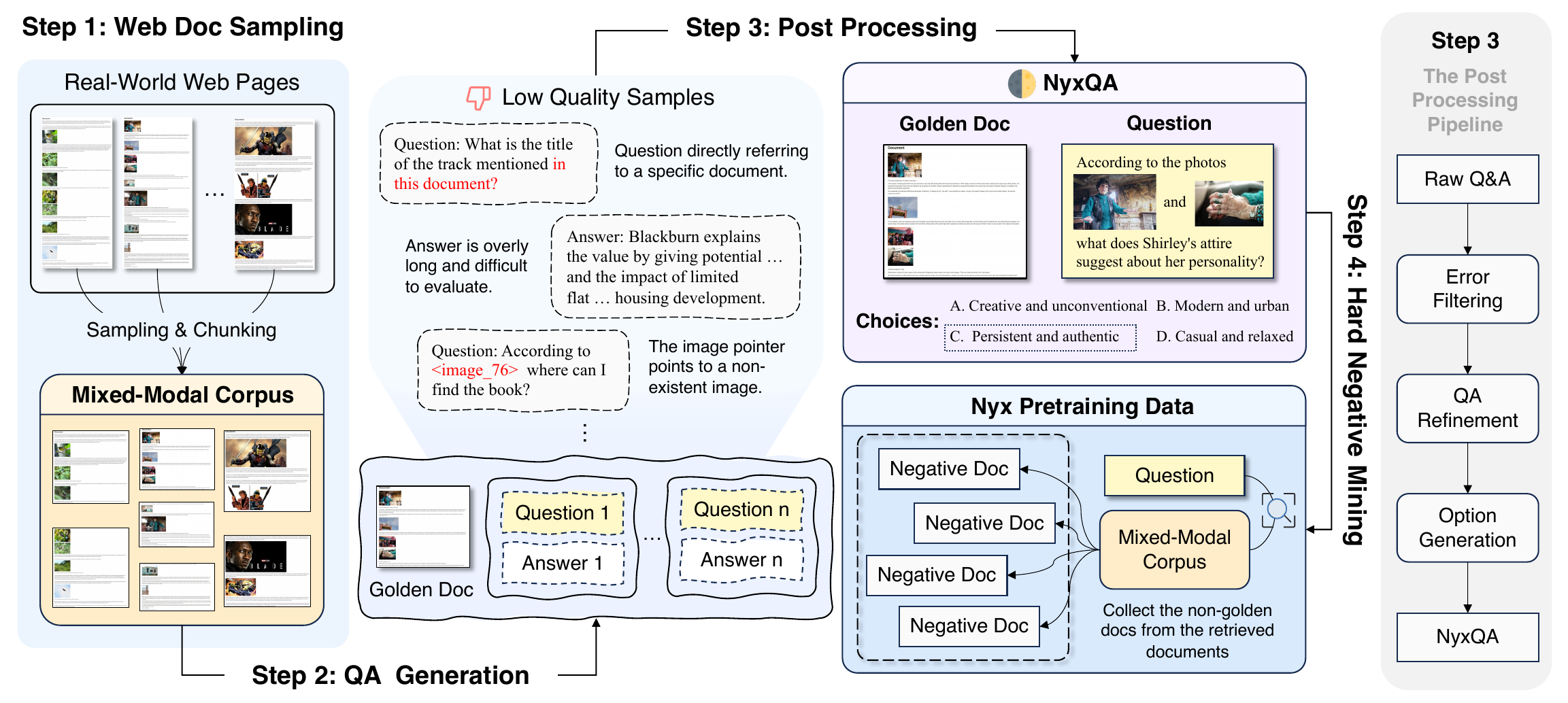}
    \caption{The proposed four-step automated \nyxqa construction pipeline.}
    \Description{}
    \label{fig:nyxqa}
\end{figure*}

\section{Related Work}
\label{sec:related_work}

\paragraph{\textbf{Multimodal Retrieval-Augmented Generation (MRAG)}} extends the traditional RAG framework to multimodal settings by retrieving text, images, or image-text pairs from an external corpus to support Vision-Language Models (VLMs) in generating textual responses~\cite{chen2022murag}. Current methods use various retrieval strategies: dual-path strategies retrieve text with text queries and images with image queries~\cite{dong2025armcts, riedler2024beyondtext}; cross-modal retrieval techniques~\cite{yu2025visrag, chen2024ragvl}; and treating multimodal documents as images for retrieval~\cite{tanaka2025vdocrag}.

In real-world applications, queries and corpora often contain \textit{mixed-modal} inputs—combinations of text and images. New multimodal deep search paradigms introduce iterative retrieval~\cite{geng2025webwatcher,wu2025mmsearch-r1,Search-o1,Li2025WebThinker}, where intermediate queries may also be mixed-modal. However, a unified retrieval framework for URAG scenarios remains undeveloped.

\paragraph{\textbf{Multimodal Embedding Retrievers}} focus on retrieving relevant multimodal documents by encoding both queries and documents into a shared embedding space. In text-only contexts, embedding-based retrievers have shown strong performance across various tasks and languages~\cite{chen2024bgem3, wang2022e5, karpukhin2020dpr,autoif,izacard2022contriever}. Extending this to multimodal scenarios, cross-modal retrievers like CLIP~\cite{radford2021clip, koukounas2024jinaclip, yuan2021florence, jia2021align} and vision-language models such as BLIP-2~\cite{li2023blip2, faysse2025colpali} encode text and images into a unified space, enabling retrieval tasks such as image-to-text, text-to-image, and image-to-image.

Recent advancements~\cite{jiang2024e5v, zhou2024vista, wei2024uniir, jiang2025vlm2vec} have leveraged VLMs as general-purpose encoders for text, images, and image-text pairs. Other studies have focused on using synthetic data~\cite{chen2025mme5, zhang2025gme, zhou2025megapairs} and improving contrastive learning objectives~\cite{lan2025llave, thirukovalluru2025b3} to enhance embedding quality. Building on this, MME~\cite{zhang2025tiir} utilized synthetic data to improve performance on interleaved text-image retrieval in the wikiHow task. However, these methods lack support for text-to-text and general interleaved text-image retrieval in URAG scenarios. Moreover, most retrievers are trained independently of downstream VLMs, resulting in suboptimal alignment. Therefore, this paper proposes a unified retriever, \nyx, which builds a bridge for mixed-modal to mixed-modal retrieval, leading to better alignment with VLM generation.

\section{Methodology}
\label{sec:method}
\subsection{Problem Formulation: URAG}

This work addresses the task of \textbf{Universal Retrieval-Augmented Generation (URAG)}, which aims to generate high-quality textual responses to mixed-modal queries by retrieving and leveraging relevant information from a mixed-modal corpus. A \textbf{mixed-modal content} $x$ is defined as an ordered sequence of elements, where each element can be either a textual segment or an image. Formally, it can be represented as: $ x \in \{ a_1 a_2 \ldots a_n\ |\ a_i \in \{ \mathcal{T}, \mathcal{I} \}\}$.

To effectively accomplish this task, we presuppose access to a \textbf{mixed-modal corpus} $\mathcal{C}$. A retriever is required to retrieve a relevant subset of \textbf{documents} $\mathcal{R}(q) \subseteq \mathcal{C}$, conditioned on a mixed-modal \textbf{question} $q$. The retrieved content then serves to guide the generation of the ultimate textual response. Formally, the objective is to learn a retrieval function $\mathcal{R}$ and a generation model $\mathcal{G}$ such that: $y = \mathcal{G}(p, q, \mathcal{R}(q))$,
where $p$ denotes a textual prompt, $q$ is the mixed-modal query, and $\mathcal{R}(q)$ represents the set of retrieved documents. The output $y$ is a \textbf{natural-language textual response}. The retrieval function $\mathcal{R}$ selects the top-$K$ most relevant entries from the corpus $\mathcal{C}$ based on their relevance to the query $q$:
\[
\mathcal{R}(q) = \mathrm{TopK}_{d \in \mathcal{C}} \, \text{sim}(q, d),
\]
where $d$ denotes a document in the mixed-modal corpus and $\text{sim}(\cdot, \cdot)$ is a similarity-based relevance function defined within the joint vision-language embedding space.

The generation model $\mathcal{G}$, typically instantiated as a VLM, processes both the query $q$ and the retrieved documents $\mathcal{R}(q)$ to yield a coherent and factually grounded textual output $y$.

\subsection{\nyxqa: A Dataset for URAG}
\label{subsec:nyxqa}

To simulate a realistic web environment, we introduce \nyxqa, a large-scale mixed-modal dataset designed for the URAG setting. Our dataset comprises three components: 
(1) a high-quality multiple-choice QA dataset $\mathcal{D}_{\text{\textsc{NyxQA}}}$ with mixed-modal questions, 
(2) a corresponding mixed-modal document corpus $\mathcal{C}_{\text{mix}}$, and 
(3) a contrastive pretraining set $\mathcal{D}_{\text{pretrain}}$ containing positive and hard negative examples for retriever training.

The construction of \nyxqa follows a four-stage pipeline. First, we sample and segment web documents to create a diverse mixed-modal corpus. Next, we utilize a VLM to generate QA pairs from these document segments. This is followed by a post-processing pipeline to filter errors, refine answers, and format multiple-choice options. Finally, we employ hard negative mining using an existing retriever to produce high-quality contrastive training triplets for pretraining the \nyx model.

\begin{figure*}[ht]
    \centering
    \includegraphics[width=\textwidth]{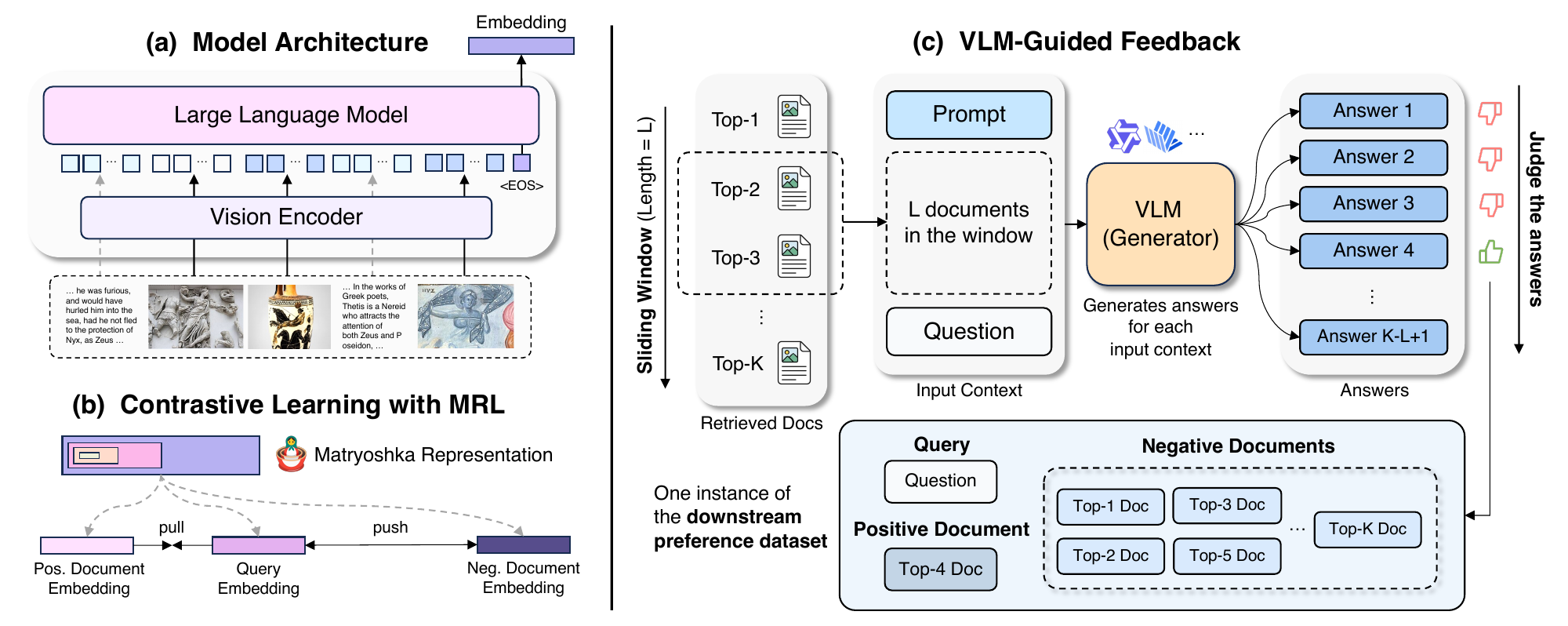}
    \caption{Overview of the \nyx architecture and its training paradigm.}
    \Description{}
    \label{fig:training}
\end{figure*}

\paragraph{\textbf{Web Document Sampling}}
To obtain naturally occurring mixed-modal documents, we sample from OBELICS~\cite{laurencon2023obelics}, a large-scale dataset of web pages featuring interleaved text and images that reflect real-world multimodal distributions. Following standard practices in text-only RAG~\cite{jin2025flashrag}, each document is segmented into smaller \textit{chunks} $\{d_i\}_{i=1}^{N}$, where each $d_i$ contains up to 200 textual tokens (excluding image tokens from the count). This segmentation maintains semantic coherence and prevents length imbalance caused by densely illustrated documents. The resulting set of chunks forms our mixed-modal corpus $\mathcal{C}_{\text{mix}} = \{d_i\}_{i=1}^{N}$, comprising 46{,}741 segments in total. We then perform stratified sampling of 10{,}000 chunks from $\mathcal{C}_{\text{mix}}$ as the basis for QA pair generation, while preserving the original modality distribution.

\paragraph{\textbf{QA Pair Generation}}
For each sampled chunk $d_i$, whether text-only or containing images, we use a VLM to generate up to five context-independent raw QA pairs $(q_{ij}^{\text{raw}}, a_{ij}^{\text{raw}})$, ensuring that each question can be answered solely based on its associated chunk. For chunks with images, we specifically prompt the VLM to create questions that reference the visual content. Since the model outputs text only, we use special tags such as \texttt{<image~k>} to denote the $k$-th image within the chunk. This process produces the raw QA dataset $\mathcal{D}_{\text{raw}} = \{(d_i, q_{ij}^{\text{raw}}, a_{ij}^{\text{raw}})\}$, which contains diverse samples ranging from pure text to multi-image questions, thereby enriching the modality diversity of \nyxqa.

\paragraph{\textbf{Post-Processing}}
The initial set $\mathcal{D}_{\text{raw}}$ of generated QA pairs is of suboptimal quality, containing various errors that could adversely affect subsequent training and evaluation. Therefore, we perform a three-stage post-processing procedure on the raw data to produce the final \nyxqa dataset.

\begin{itemize}
    \item \textbf{Error Filtering.} Questions with explicit contextual references (e.g., phrases like ``in this document'') are removed using rule-based filters. In addition, we ensure image–text consistency by verifying that the image tags mentioned in the generated question correspond to actual images present in the chunk $d_i$.
    \item \textbf{QA Refinement.} We further refine the filtered QA pairs using a VLM to enhance clarity and completeness. Each retained pair $(q_{ij}^{\text{raw}}, a_{ij}^{\text{raw}})$ is compressed to its essential content, eliminating redundancy while preserving factual accuracy. This process yields concise, self-contained questions and answers that align closely with the corresponding gold document $d_i^{+}$, resulting in the refined set $(d_i^{+}, q_{ij}, a_{ij}^{+})$.
    \item \textbf{Option Generation.} For each refined QA, an LLM generates three semantically plausible distractors $\{a_{ij}^{-}\}$ for the question $q_{ij}$. After shuffling the distractors with the correct answer, we finalize each sample with the question, four options, and the gold document, forming our multiple-choice dataset $\mathcal{D}_{\text{\textsc{NyxQA}}} = \{(q_{ij}, \{a_{ij}^{+}, a_{ij}^{-}\}, d_i^{+})\}$.
\end{itemize}

\paragraph{\textbf{Hard Negative Mining}}
To enhance retriever pretraining, we construct contrastive triplets using $\mathcal{D}_{\text{\textsc{NyxQA}}}$. Each question $q_{ij}$ serves as a query, with its corresponding gold document $d_i^{+}$ designated as the positive sample. We then employ mmE5~\cite{chen2025mme5} to retrieve the top-10 relevant documents from the mixed-modal corpus $\mathcal{C}_{\text{mix}}$. From these, we select five documents that differ from $d_i^{+}$ as hard negatives $\{d_{ij}^{-}\}$, prioritizing the highest-ranked candidates. This yields the pretraining dataset $\mathcal{D}_{\text{pretrain}} = \{(q_{ij}, d_i^{+}, \{d_{ij}^{-}\})\}$, a contrastive training set specifically designed for mixed-modal retrieval.

\subsection{\nyx: Training Paradigm}

\paragraph{\textbf{Overview}} Our goal is to build a unified retriever capable of handling mixed-modal queries and documents across diverse real-world scenarios. To this end, we begin by pretraining \nyx on a large-scale corpus that includes both public and synthetic datasets spanning various modality configurations. This initialization equips the retriever with general-purpose retrieval capabilities across text-only, image-only, and multimodal pairs.

However, generic pretraining may not fully align with the specific information needs of downstream VLMs during generation. Therefore, in the second stage, we fine-tune \nyx-pretrained through a feedback-driven learning process, leveraging VLM responses to construct high-quality examples that reflect the actual relevance signals needed for multimodal generation.

Throughout both stages, we employ contrastive learning with Ma Matryoshka Representation Learning~\cite{kusupati2022mrl} to ensure scalable and efficient embedding quality under varying dimensional constraints. We detail our training objective and the two-stage procedure below.

\paragraph{\textbf{Training Objective}}

We build our retriever on top of a pretrained VLM, Qwen-2.5-VL-3B-Instruct~\cite{tongyi2025qwen25vl}, as the backbone encoder. Given an input sequence, we use the hidden representation of the final \texttt{<EOS>} token as the global embedding for retrieval.

Following established practices in embedding model training, we construct each training instance as a triplet \( \{ q, d^+, \{ d_n^- \}_{n=1}^N \} \), where \( q \) is a query, \( d^+ \) is a positive document, and \( \{ d_n^- \} \) are \( N \) negative documents. An instruction string is prepended to each query before encoding. Both queries and documents may come from mixed modalities (e.g., text, image, or interleaved image-text), allowing \nyx to operate in a unified embedding space.

To learn discriminative representations, we adopt the InfoNCE loss for contrastive learning. Let \( \mathbf{h}_q \in \mathbb{R}^d \), \( \mathbf{h}^+ \in \mathbb{R}^d \), and \( \{ \mathbf{h}_n^- \}_{n=1}^N \subset \mathbb{R}^d \) denote the embeddings of the query, the positive document, and the negative documents, respectively. We apply Matryoshka Representation Learning (MRL)~\cite{kusupati2022mrl}, which encourages the full embedding \( \mathbf{h} \in \mathbb{R}^d \) to remain informative even when truncated to lower-dimensional subspaces. This enables flexible trade-offs between retrieval performance and memory efficiency.

Specifically, for a set of target dimensions \( \{ d_1, d_2, \ldots, d_K \} \), where \( d_k < d \), we truncate each embedding to its first \( d_k \) dimensions, denoted as \( \mathbf{h}^{(d_k)} \in \mathbb{R}^{d_k} \). For each \( d_k \), we compute an InfoNCE loss as:
\begin{equation}
\mathcal{L}_{\text{Info}}^{(d_k)} = - \log \frac{
    \phi(\mathbf{h}_q^{(d_k)}, \mathbf{h}^{+ (d_k)})
}{
    \phi(\mathbf{h}_q^{(d_k)}, \mathbf{h}^{+ (d_k)}) + \sum_{n=1}^{N} \phi(\mathbf{h}_q^{(d_k)}, \mathbf{h}_n^{- (d_k)})
},
\end{equation}

\noindent where \( \phi(\mathbf{a}, \mathbf{b}) = \exp\left( \mathrm{sim}(\mathbf{a}, \mathbf{b}) / \tau \right) \),
and \( \mathrm{sim}(\cdot, \cdot) \) denotes cosine similarity with temperature hyperparameter \( \tau > 0 \).
Here, \( \mathbf{h}^{+ (d_k)} \) and \( \mathbf{h}_n^{- (d_k)} \) correspond to the positive and the \(n\)-th negative sample document embeddings, respectively.

The final training objective aggregates the InfoNCE losses over all truncated dimensions as a weighted sum:
\begin{equation}
\mathcal{L}_{\text{MRL}} = \sum_{k=1}^{K} w_k \cdot \mathcal{L}_{\text{Info}}^{(d_k)}, \quad \text{where} \sum_{k=1}^{K} w_k = 1,
\end{equation}
\noindent with \( w_k \) denoting the weight for the \( k \)-th dimension. This objective encourages each embedding prefix to preserve semantic integrity under varying retrieval constraints.

\paragraph{\textbf{Stage 1: Pretraining with Mixed-Modal Data.}}

In the first stage, we pretrain \nyx as a general-purpose retriever using a large-scale corpus constructed from both public and synthetic data sources. Following mmE5~\cite{chen2025mme5}, we include MMEB~\cite{jiang2025vlm2vec} and synthetic triplets from the mmE5 pipeline. To ensure the model's ability to handle genuinely mixed-modal scenarios, we further integrate our proposed \nyxqa dataset, which supports retrieval across diverse modality combinations.

Since real-world retrieval tasks still predominantly involve textual inputs, we further enhance the retriever's text understanding ability by introducing additional text-only datasets. Specifically, we use the training sets from HotpotQA~\cite{yang2018hq}, 2WikiMultiHopQA~\cite{ho20202wiki}, and MuSiQue~\cite{trivedi2022musique}. For each query, we retrieve the top-\( K \) documents from the full Wikipedia corpus using E5-v2~\cite{wang2022e5}, and treat the top-1 document as the positive sample, while selecting negative samples from documents ranked beyond top-10.

All the datasets described above are combined and jointly used to train the initial retriever, resulting in the \nyx-pretrained model.

\paragraph{\textbf{Stage 2: Supervised Fine-tuning with VLM-Guided Feedback.}}

While \nyx-pretrained demonstrates strong retrieval performance, it is not explicitly optimized for supporting downstream generation by VLMs. To bridge this gap, we introduce a fine-tuning stage that leverages feedback from a VLM to align the retriever with the actual information needs during generation.

Given a dataset \( D = \{(q_i, a_i)\} \) of queries and their corresponding answers, along with a retrieval corpus \( \mathcal{C} \), we proceed as follows. For each query \( q_i \), we first use \nyx-pretrained to retrieve the top-\( K \) candidate documents \( \{d_1, d_2, \ldots, d_K\} \). Then, using a sliding window of length \( L \), we construct a sequence of candidate contexts by grouping contiguous subsets of the retrieved documents. Each context window is concatenated with the query and fed into the VLM to generate an answer.

We select the first context window that either yields a correct answer or exceeds a pre-defined generation metric threshold (e.g. EM, F1). The first document in this window is treated as the positive sample \( d^+ \), and the remaining \( K - 1 \) documents are used as negative samples \( \{d_n^-\} \). If no window meets the quality threshold, the entire instance is discarded from the feedback dataset.

By applying this procedure to all queries in \( D \), we construct a \textbf{downstream preference dataset} from the feedback \( D_{\text{pref}} = \{(q_i, d_i^+, \{d_{i,n}^-\})\} \), which reflects the actual preferences of the VLM in real generation scenarios. We then fine-tune \nyx-pretrained on this dataset using the same contrastive learning framework described earlier, thus obtaining the final retriever \nyx.

\begin{table*}[ht]
    \centering
    \caption{The overall results on the six RAG datasets. To ensure consistent evaluation, the top document retrieved by each retriever was combined with the corresponding question, then input into Qwen2.5-VL-7B for answer generation. The exception is SciQA, where the retrieval content consists of one lecture and two example-based retrieval results to suit the dataset's structure. This setup isolates the effect of the retrievers, facilitating a controlled comparison of retrieval performance. The best results are highlighted in bold, and the second-best results are underlined.}
    \vspace{-1em}
    \begin{tabularx}{\textwidth}{l *{6}{Y} | Y *{3}{Y} | Y Y}
    \toprule
    \multirow{2}{*}{\textbf{Method}}
     & \multicolumn{2}{c}{\textbf{HotpotQA}}
     & \multicolumn{2}{c}{\textbf{Bamboogle}}
     & \multicolumn{2}{c}{\textbf{MuSiQue}}
     & \multicolumn{1}{|c}{\textbf{SciQA}}
     & \multicolumn{2}{c}{\textbf{MMQA}}
     & \textbf{\nyxqa}
     & \multirow{2}{*}{\textbf{Avg.}} \\
    \cmidrule(lr){2-3}
    \cmidrule(lr){4-5}
    \cmidrule(lr){6-7}
    \cmidrule(lr){8-8}
    \cmidrule(lr){9-10}
    \cmidrule(lr){11-11}
     & EM & F1 & EM & F1 & EM & F1 & Acc & EM & F1 & Acc &  \\
    \midrule
    
    \multicolumn{12}{l}{\textbf{\textit{Direct Answer}}} \\
    InternVL3 (8B)~\cite{zhu2025internvl3} & 16.40 & 22.88 & 9.60 & 15.49 & 3.60 & 8.16 & 78.87 & 20.07 & 23.99 & 53.33 & 25.31 \\
    Qwen2.5-VL (7B)~\cite{tongyi2025qwen25vl} & 12.40 & 18.36 & 6.40 & 11.50 & 3.29 & 7.32 & 77.98 & 20.73 & 24.39 & 50.17 & 24.38 \\
    \midrule
    
    \multicolumn{12}{l}{\textbf{\textit{Text RAG}}} \\
    E5-v2 (109M)~\cite{wang2022e5} & 14.40 & 19.18 & 7.20 & 12.80 & 2.40 & 6.79 & -- & -- & -- & -- & -- \\
    \midrule
    
    \multicolumn{12}{l}{\textbf{\textit{Vision-Language RAG}}} \\
    CLIP (150M)~\cite{radford2021clip} & 14.00 & 21.12 & 6.40 & 11.64 & 3.20 & 6.74 & 73.07 & 18.03 & 20.67 & 61.50 & 23.64 \\
    VLM2Vec (4B)~\cite{jiang2025vlm2vec} & 14.40 & 22.08 & 10.40 & 16.95 & 3.60 & 10.12 & 79.56 & 19.91 & 23.34 & 56.50 & 25.69 \\
    VisRAG-Ret (3B)~\cite{yu2025visrag} & 12.08 & 19.84 & 8.80 & 16.05 & 3.60 & 8.29 & 80.45 & 18.84 & 21.55 & 64.33 & 25.38 \\
    mmE5 (11B)~\cite{chen2025mme5} & 17.60 & 24.30 & 13.60 & 18.69 & 5.20 & 9.70 & \underline{81.40} & \underline{34.00} & \underline{38.50} & 66.83 & 30.98 \\
    \midrule
    
    \multicolumn{12}{l}{\textbf{\textit{Ours}}} \\
    \nyx-pretrained (3B) & \underline{22.00} & \underline{31.38} & \underline{16.00} & \underline{22.87} & \underline{5.60} & \underline{11.00} & 81.33 & 31.75 & 35.97 & \underline{74.83} & \underline{33.27} \\
    \nyx (3B) & \textbf{24.40} & \textbf{33.19} & \textbf{16.80} & \textbf{25.93} & \textbf{7.20} & \textbf{12.80} & \textbf{81.75} & \textbf{39.66} & \textbf{44.50} & \textbf{81.83} & \textbf{36.46} \\
    \bottomrule
    \end{tabularx}
    \label{tab:main_results}
\end{table*}

\section{Main Experiments}
\label{sec:experiments}

Our main experiments consist of two parts, where we first evaluate the generation performance in URAG scenarios and then examine the embedding performance, given that the model is inherently an embedding model. All experiments were conducted on a single node equipped with 8 $\times$ \texttt{NVIDIA A800-SXM4-80GB} GPUs. For efficient training, we applied LoRA~\cite{hu2022lora} with a rank of 8. The per-device batch size was set to 20 with 4 gradient accumulation steps, and the temperature parameter $\tau$ in the InfoNCE loss was fixed at 0.02. To avoid memory overflow when processing multi-image samples, the maximum visual input resolution was limited to \(400 \times 28 \times 28\) pixels. Additional implementation details can be found in the appendix.

\subsection{Experimental Setup}

\paragraph{\textbf{Datasets and Metrics}} We evaluate RAG pipelines incorporating our retriever across two categories: (1) text-only datasets and (2) multimodal datasets, including MRAG and URAG. 

For text-only RAG, following \textsc{ReCall}~\cite{chen2025recall}, we evaluate on \textbf{HotpotQA}~\cite{yang2018hq}, \textbf{MuSiQue}~\cite{trivedi2022musique}, and \textbf{Bamboogle}~\cite{press2023bamboogle}. For each question, we use E5-v2 to retrieve the top 20 documents from Wikipedia, merge and deduplicate them to create a task-specific corpus, and randomly sample up to 250 questions per dataset.

In the case of MRAG and URAG, we utilize \textbf{MultimodalQA} (MMQA)~\cite{talmor2021mmqa}, \textbf{ScienceQA} (SciQA)~\cite{lu2022sciqa}, and \nyxqa. MMQA requires retrieval from a corpus containing text, tables, and images. Since the tables in the MMQA corpus are originally stored in JSON format, we employ the Python \texttt{tabulate} library to convert them into a more comprehensible text table format. SciQA presents image-text questions, and we construct its corpus using associated lectures and QA examples. \nyxqa, constructed from authentic web pages, covers a broader spectrum of input and document modalities, thereby facilitating more realistic and comprehensive evaluation of URAG in real-world web environments.

During pretraining, we use the training sets of 2WikiMultiHopQA, HotpotQA, MuSiQue, and \nyxqa, treating Bamboogle, MMQA, and SciQA as out-of-domain (OOD) evaluation sets. For feedback-based fine-tuning, feedback is collected from HotpotQA, MuSiQue, MMQA, SciQA, and \nyxqa, with Bamboogle serving as the OOD benchmark.

For multiple-choice datasets, we report \textbf{accuracy} (Acc), while for open-ended QA, we adopt \textbf{exact match} (EM) and \textbf{F1 score} (F1), reflecting both strict correctness and token-level overlap between predictions and references.

\paragraph{\textbf{Baseline Models.}} For the text-only retriever, we use E5-v2~\cite{wang2022e5} as the unimodal RAG baseline, since it serves as the backbone model for constructing our text-only retrieval datasets and is also one of the most widely used retrievers in text-based RAG systems~\cite{jin2025flashrag}.
For multimodal retrievers, we use well-supervised fine-tuned embedding models CLIP~\cite{radford2021clip}, VLM2Vec~\cite{jiang2025vlm2vec} and mmE5~\cite{chen2025mme5}, as well as a retriever for visual document retrieval for RAG, VisRAG-Ret~\cite{yu2025visrag}. We also report the direct answering results of InternVL3-8B~\cite{zhu2025internvl3} and Qwen2.5-VL-7B as baselines for comparison.

\subsection{Results on Generation Performance}
Our generation performance results are presented in Table~\ref{tab:main_results}. Overall, \nyx consistently outperforms all baselines, clearly demonstrating its superiority. We further highlight the following insights:

\paragraph{\textbf{Performance in Text-Only RAG}} 
Despite the powerful 11 billion parameter VLM backbone of mmE5, our 3 billion parameter \nyx-pretrained model still outperforms mmE5 on HotpotQA, Bamboogle, and MuSiQue, with performance gains of 9\% and 6\% on HotpotQA and Bamboogle, respectively. This result shows the strength of targeted training. Moreover, \nyx substantially surpasses the text-only retriever E5 that is commonly used in RAG frameworks, further demonstrating its effectiveness in unimodal retrieval.

\paragraph{\textbf{Multimodal RAG Performance}} 
In multimodal tasks, \nyx-pretrained performs competitively on MMQA and \nyxqa, though it trails mmE5 slightly on SciQA. This may be attributed to \nyx's smaller parameter count and its broader training coverage, which includes interleaved and text-only examples. Nevertheless, its robust performance across different input types highlights the benefit of mixed-modal training.
After incorporating feedback from downstream VLMs, \nyx achieves the best performance across all multimodal benchmarks, with great results on MMQA (F1: 35.97\% $\rightarrow$ 44.50\%) and \nyxqa (Accuracy: 74.83\% $\rightarrow$ 81.83\%). On SciQA, the gain is modest, possibly due to the limited informativeness of the provided lecture corpus. Nonetheless, fine-tuning with feedback still leads to alignment with the VLM's preferences. 

A McNemar’s test was conducted on \nyxqa to assess the performance differences among mmE5, \nyx-pretrained, and \nyx as retrievers. 
The comparison between mmE5 and \nyx-pretrained yielded a test statistic of 19.0631 ($\chi^{2}$, 1 degree of freedom) with a p-value of 0.0000. 
Furthermore, the comparison between \nyx-pretrained and \nyx resulted in a test statistic of 15.7538 and a p-value of 0.0001. 
These results provide strong evidence that the retrieval performance differs significantly across the methods.

\paragraph{\textbf{Beyond Gold Documents: Learning from Preference}} 
An interesting observation arises from \nyxqa, where each question is originally paired with a generation-originated ``golden'' document. Although semantically relevant, these gold documents do not always lead to correct answers during inference. Our feedback analysis shows that documents preferred by the VLM may differ from the labelled positives. Incorporating this preference signal during fine-tuning leads to a 7-point accuracy gain on \nyxqa. This suggests the importance of further aligning retrieval models with downstream generative utility in URAG systems. 

\begin{table}[!ht]
    \centering
    \caption{Performance comparison on the MMEB benchmark, which includes 36 tasks spanning four categories: classification (Class.), visual question answering (VQA), retrieval (Retr.), and visual grounding (Ground.).}
    \vspace{-1em}
    \begin{tabular}{lccccc}
    \toprule
    \multirow{2}{*}{\textbf{Models}} & \multicolumn{4}{c}{\textbf{Per Meta-Task Score}} & \multirow{2}{*}{\textbf{Overall}} \\
    \cmidrule(lr){2-5}
     & Class. & VQA & Retr. & Ground. \\
    \midrule
    CLIP~\cite{radford2021clip}       & 42.8 &  9.1 & 53.0 & 51.8 & 37.8 \\
    BLIP2~\cite{li2023blip2}      & 27.0 &  4.2 & 33.9 & 47.0 & 25.2 \\
    OpenCLIP~\cite{cherti2023openclip}   & 47.8 & 10.9 & 52.3 & 53.3 & 39.7 \\
    E5-V~\cite{jiang2024e5v}       & 21.8 &  4.9 & 11.5 & 19.0 & 13.3 \\
    MagicLens~\cite{zhang2024magiclens}  & 38.8 &  8.3 & 35.4 & 26.0 & 27.8 \\
    MMRet~\cite{zhou2025megapairs}      & 47.2 & 18.4 & 56.5 & 62.2 & 44.0 \\
    VLM2Vec~\cite{jiang2025vlm2vec} & 52.8 & 50.3 & 57.8 & 72.3 & 55.9\\    
    mmE5~\cite{chen2025mme5}       & 67.6 & 62.8 & 70.9 & 89.7 & 69.8 \\
    mmE5-Qwen-3B & 56.6 & 56.0 & 59.4 & 71.5 & 59.0 \\
    \midrule
    \nyx-pretrained & 55.2 & 53.7 & 58.4 & 70.5 & 57.5 \\ 
    \nyx & 57.9 & 57.5 & 61.8 & 75.7 & 61.1 \\
    \bottomrule
    \end{tabular}
    \vspace{-1em}
    \label{tab:mmeb}
\end{table}

\subsection{Embedding Capability Analysis}
Although aligning mixed-modal retriever with downstream models can enhance the generative quality of the final VLM, the capability of the retriever itself is also of central importance. We evaluate the embedding ability of our models on the MMEB benchmark~\cite{jiang2025vlm2vec}, and the results are reported in Table~\ref{tab:mmeb}. In the table, models from CLIP to MMRet are evaluated in the zero-shot setting, whereas the remaining models are trained with MMEB-labelled data. In particular, \textit{mmE5-Qwen2.5-3B} is trained on Qwen-2.5-VL-3B-Instruct, with its training data consisting of MMEB-labelled data together with the retrieval and VQA subsets of the mmE5 synthetic data, serving as the ablation setting.  

Compared to mmE5, \textit{mmE5-Qwen2.5-3B} performs worse across all capabilities, which can be attributed to its smaller backbone size and the exclusion of the classification subset from the mmE5 synthetic data (to maintain alignment with \nyx-pretrained and \nyx settings). Nevertheless, it still surpasses other baseline models. When OOD pure text data and \nyxqa mixed-modal data are included, the mismatch with the MMEB evaluation pattern results in a 1.5\% overall performance drop. However, after fine-tuning with feedback from a VLM on OOD datasets with entirely different tasks, \nyx outperforms \textit{mmE5-Qwen2.5-3B} across all capabilities, achieving a 2.1\% overall improvement. These findings further demonstrate that incorporating VLM feedback not only improves the performance of URAG systems but also enhances the embedding capability of dense retrievers themselves.

\section{Quantitative Analysis}

\subsection{Impact of Data Scale on URAG Performance}

The scalability of training data is crucial for building effective retrievers. Prior studies have shown an approximately logarithmic-linear relationship between the volume of training data and the quality of retrieval model embeddings~\cite{chen2025mme5, fang2024scaling}. In this section, we further examine how the scale of training data affects \nyx's performance in the URAG setting. 

\begin{figure}[!t]
    \centering
    \includegraphics[width=0.45\textwidth]{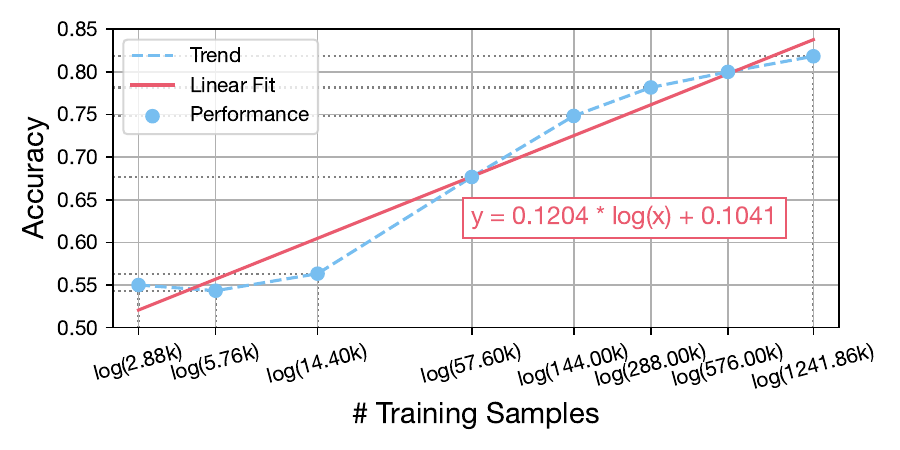}
    \vspace{-1em}
    \caption{Impact of training data scale on \nyxqa accuracy when training \nyx with varying sample sizes.}
    \Description{}
    \label{fig:data_scale}
\end{figure}

As illustrated in Figure~\ref{fig:data_scale}, the performance trend closely follows a logarithmic-linear curve, consistent with previous findings. The steady improvement of URAG performance with increasing data scale further confirms the high quality and diversity of our training data. This indicates that enhancements in the retriever's independent capabilities translate proportionally into gains in end-to-end URAG performance. Thus, increasing training data is expected to predictably enhance URAG scenario generalization.

\subsection{Effect of Retrieved Document Count}

\begin{figure}[!t]
    \centering
    \includegraphics[width=0.45\textwidth]{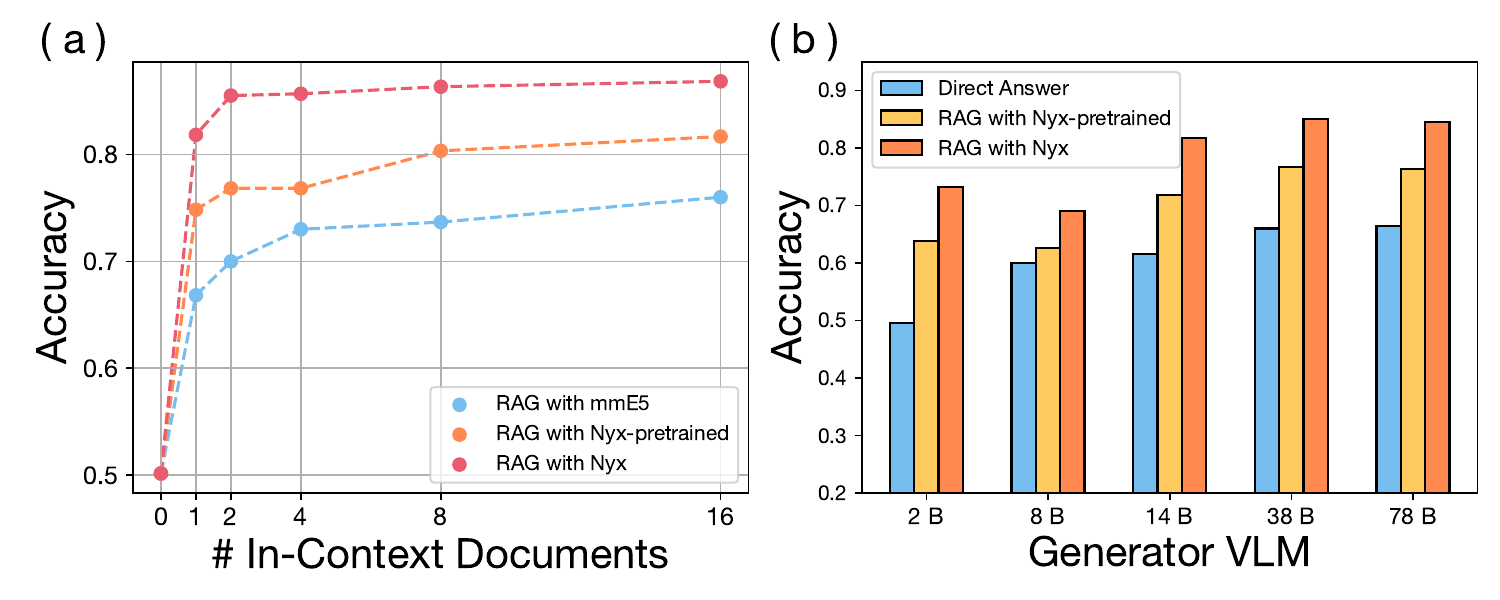}
    \vspace{-1em}
    \caption{Impact of (a) the number of in-context documents and (b) feedback-based retriever fine-tuning on downstream generation performance. Results are shown on \nyxqa using InternVL3 models of varying sizes, respectively.}
    \Description{}
    \label{fig:gen_icl}
\end{figure}

To examine how the number of retrieved documents influences generation quality, we vary the number of documents fed into Qwen2.5-VL-7B from 0 to 16, evaluating the URAG results of \textbf{mmE5}, \nyx-pretrained, and \nyx. As shown in Figure~\ref{fig:gen_icl}(a), adding more documents consistently improves all retrievers, though the gains diminish as the count increases. \nyx consistently outperforms both \nyx-pretrained and mmE5, demonstrating robust performance even with fewer documents and confirming the effectiveness of feedback-based fine-tuning in producing informative retrievals. Overall, the results highlight the critical importance of high-quality top-ranked retrieval for efficient generation.


\begin{figure*}[th]
    \centering
    \includegraphics[width=0.95\textwidth]{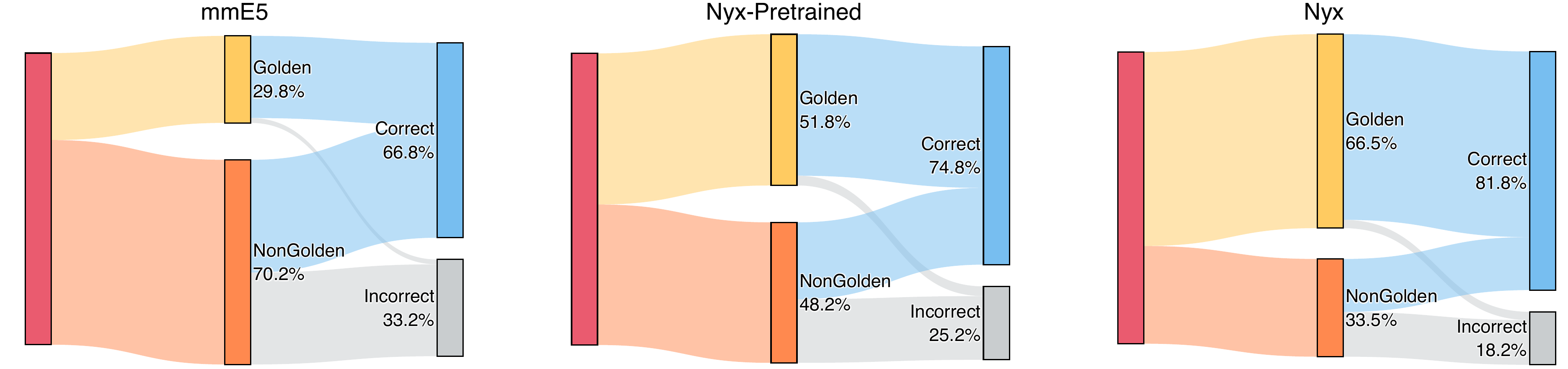}

    \caption{Comparison of retrieval and answer correctness distributions on \nyxqa for mmE5, \nyx-pretrained, and \nyx.}
    \Description{}
    \label{fig:sankey}
\end{figure*}

\begin{figure}[th]
    \centering
    \includegraphics[width=0.45\textwidth]{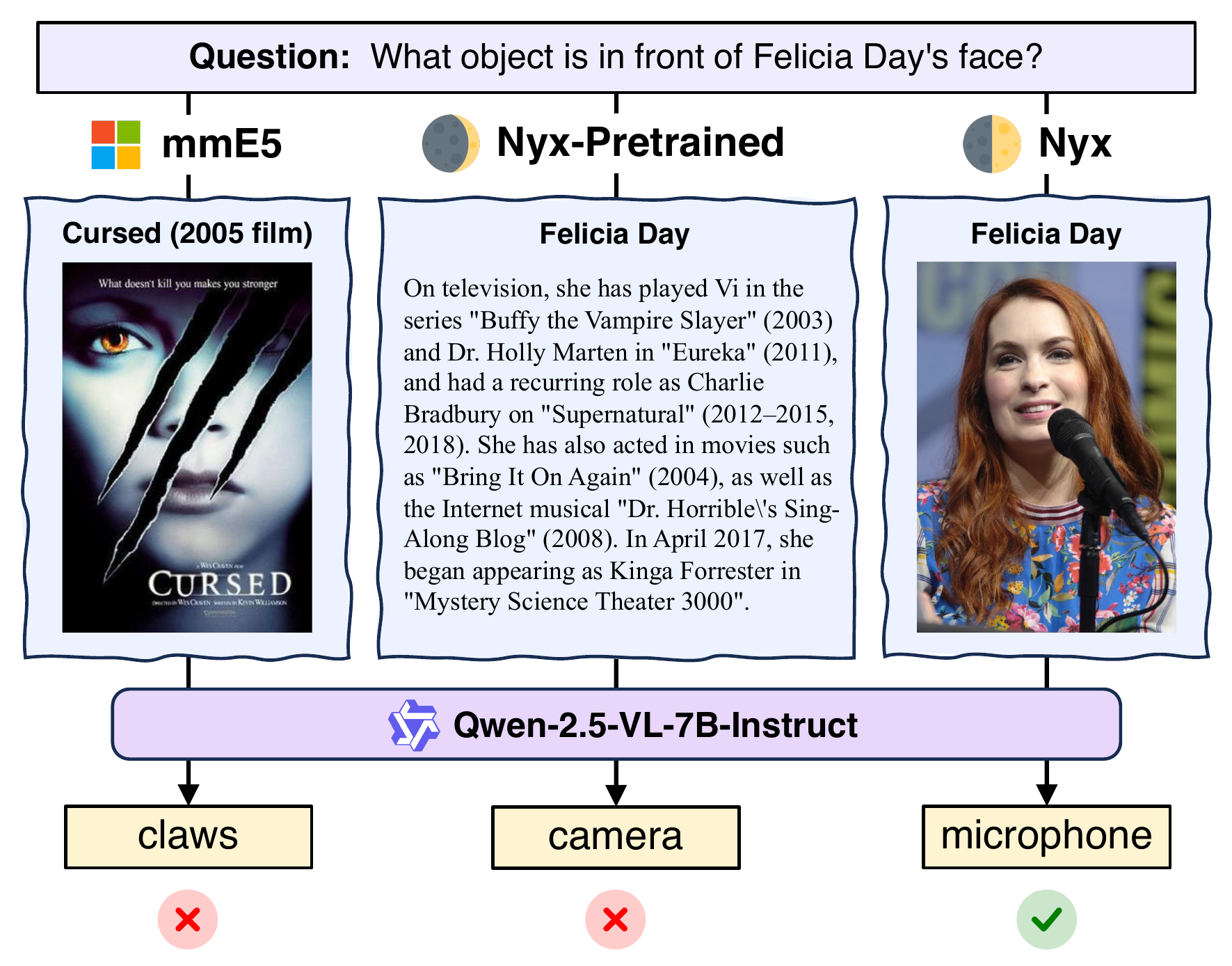}
    \caption{Case study on MMQA. The top-1 retrieved documents by mmE5, \nyx-pretrained, and \nyx are shown together with the corresponding answers produced by VLM.}
    \Description{}
    \label{fig:case_study}
\end{figure}

\subsection{Generalization Across Generators}

While \nyx is fine-tuned with supervision from Qwen2.5-VL-7B, we also examine whether such supervision generalizes to other VLMs. To assess this, we evaluate its performance across InternVL3 models of varying sizes used as generators. As shown in Figure~\ref{fig:gen_icl} (b), \nyx consistently outperforms the direct-answer baseline across all InternVL3 variants, indicating that supervision from Qwen2.5-VL-7B transfers effectively across different generator architectures. Integrating \nyx yields further improvements, particularly for InternVL3-2B and InternVL3-14B, with absolute gains exceeding 0.2 points.

However, the degree of improvement varies with generator size, indicating different alignment preferences among models. Additionally, since performance does not increase monotonically with model size, this suggests that generator size is not a reliable predictor of RAG pipeline performance. Effective alignment is crucial in bridging semantic gaps across VLMs.

\subsection{Effect of MRL}

In real-world retrieval systems, reducing embedding dimensions can significantly decrease memory usage and speed up retrieval. To adapt to varying resource budgets, we incorporate MRL into our contrastive training framework. MRL ensures the model maintains meaningful representations across reduced dimensions. We train the model to generate effective embeddings at four target dimensions: 2048, 1024, 512, and 256, with weights of [1.0, 1.0, 0.2, 0.2] (before normalization), where 2048 is the default VLM output.

As shown in Table~\ref{tab:mrl}, the 1024-dimensional variant achieves accuracy comparable to the 2048 one while halving storage, and even the 512- and 256-dimensional versions maintain strong performance. These results highlight MRL’s ability to provide efficient, resource-aware retrieval with graceful performance degradation under limited memory or latency budgets.

\subsection{Impact of Retrieved Docs on Generation}
To investigate the impact of retrieved documents on generation, we begin with a case study on MMQA. As shown in Figure~\ref{fig:case_study}, we examines how retrieval influences the produced answers. Unlike MMEB, which assumes modality-specific similarity computation, real-world retrieval entails cross-modal relevance estimation across multimodal documents. In this example, mmE5 retrieves an image-text pair focused solely on ``face,'' missing the query subject; \nyx-pretrained correctly identifies ``Felicia Day'' but provides the textual evidence that fails to support the answer; in contrast, \nyx retrieves the correct entity along with both the proper title and visual information, directly grounding the generated response. 
\begin{table}[!t]
    \centering
    \renewcommand{\arraystretch}{1.2}
    \caption{Performance of \nyx on \nyxqa under different output dimensions}

    \begin{tabular}{lcccc}
    \toprule
    \multirow{2}{*}{\textbf{Setting}}
    & \multicolumn{4}{c}{\textbf{Output Embedding Dimension}} \\
    \cmidrule(lr){2-5}
    & \textbf{2048-dim} & \textbf{1024-dim} & \textbf{512-dim} & \textbf{256-dim} \\
    \midrule
    \textbf{Weight}   & 1.0000 & 1.0000 & 0.2000 & 0.2000 \\
    \textbf{Accuracy} & 0.8183 & 0.8100 & 0.7800 & 0.7467 \\
    \bottomrule
    \end{tabular}
    \label{tab:mrl}
\end{table}

Building on these qualitative observations, we further perform a quantitative analysis on the \nyxqa dataset to study the relationship between retrieval correctness and answer correctness. The retrieval quality is visualized in Figure~\ref{fig:sankey} using Sankey diagrams. The results reveal two key trends: (1) higher proportions of golden documents lead to higher answer accuracy; and (2) even with non-golden documents, nearly half of the answers remain correct, demonstrating the robustness of VLMs. These findings suggest that improving retrievers is crucial not only for ensuring faithful grounding but also for mitigating noise from irrelevant evidence. Future improvements may arise from modelling VLM preferences on non-golden evidence, which can sometimes diverge from human intuition.

\section{Conclusion}
To enable Universal Retrieval-Augmented Generation (URAG) over arbitrarily mixed-modal questions and corpora, we constructed \nyxqa, the first large-scale and comprehensive dataset that faithfully reflected real-world URAG scenarios, where text, images, and their interleaved combinations naturally coexisted. Building on this foundation, we introduced \nyx, a unified multimodal retriever explicitly optimized for such settings. \nyx was initially pretrained via contrastive learning with Matryoshka Representation Learning on a diverse mixture of public and synthetic data, and was subsequently fine-tuned using feedback from a downstream vision-language generator, thereby better aligning retrieval relevance with generation utility. Extensive experiments demonstrated that this simple yet effective pipeline achieved consistent and substantial improvements over both unimodal and multimodal baselines across all modality combinations, underscoring the promise of unified mixed-modal retrieval for next-generation URAG systems.

\bibliographystyle{ACM-Reference-Format}
\bibliography{references}

\input{appendix}

\end{document}

%% file: appendix.tex
\newtcolorbox{promptbox}[1][]{
  colback=gray!5!white,
  colframe=gray!75!black,
  fonttitle=\bfseries,
  left=5pt,
  right=5pt,
  top=5pt,
  bottom=5pt,
  breakable,
  enhanced,
  #1
}

\clearpage
\section*{Appendix}
\appendix
\section{Training Details}
We train Nyx based on the Qwen2.5-VL-3B model using a single node equipped with 8$\times$NVIDIA A800-SXM4-80GB GPUs. To enable efficient fine-tuning, we apply Low-Rank Adaptation (LoRA)~\cite{hu2022lora} with a rank of 8. Each GPU processes a batch of 20 samples, and we accumulate gradients over 4 steps, resulting in an effective batch size of 640. To prevent memory overflow when processing multi-image inputs, we cap the visual input resolution at 400$\times$28$\times$28 pixels.

We use DeepSpeed with bf16 mixed-precision training and enable gradient checkpointing for memory efficiency. The optimizer is AdamW, combined with a linear learning rate scheduler. The base learning rate is set to 1e-5, with a warmup ratio of 0.05, and a maximum gradient norm of 5.0. The contrastive loss function uses a temperature of 0.02 and a negative sampling ratio of 1.

For efficient memory optimization, we employ DeepSpeed's ZeRO optimization at stage 2, which partitions model states across devices to significantly reduce memory consumption. This allows us to train larger models without overflow. ZeRO stage 2 also enables communication optimizations such as allgather and reduce-scatter, improving parallelism and computational efficiency. These configurations, combined with gradient checkpointing and mixed-precision training, work together to maximize both performance and memory efficiency during training.

\section{Baseline Retriever Models}
In this section, we present the baseline retriever models employed in our main experiments. These models capture several recent advances in text and multimodal retrieval, and provide strong baselines for assessing the effectiveness of our proposed method.

\smallskip
\noindent\textbf{E5}~\cite{wang2022e5} is a series of cutting-edge text embeddings that perform exceptionally well across a variety of tasks. The model is trained using a contrastive approach, with weak supervision signals derived from a carefully curated large-scale text pair dataset. It demonstrates strong performance both in zero-shot scenarios and after fine-tuning.

\smallskip
\noindent\textbf{CLIP}~\cite{radford2021clip} is a powerful multimodal model developed by OpenAI that learns visual and textual representations jointly. It is trained using a large dataset of image-text pairs in a contrastive manner, enabling it to understand images and texts in a shared embedding space. This approach allows CLIP to perform a variety of tasks, such as zero-shot image classification, image search, and text-to-image retrieval, without task-specific fine-tuning.

\smallskip
\noindent\textbf{VLM2Vec}~\cite{jiang2025vlm2vec} is a versatile multimodal embedding model designed to convert any state-of-the-art VLM into a unified embedding space. It employs a contrastive training framework on the Massive Multimodal Embedding Benchmark (MMEB), which encompasses four meta-tasks—classification, visual question answering, multimodal retrieval, and visual grounding—across 36 datasets. Unlike models such as CLIP or BLIP, which process text and images independently, VLM2Vec integrates both modalities based on task-specific instructions to produce fixed-dimensional vector representations.

\smallskip
\noindent\textbf{mmE5}~\cite{chen2025mme5} is a multimodal multilingual embedding model that enhances performance by leveraging high-quality synthetic datasets. These datasets encompass a wide range of tasks, modality combinations, and languages, and are generated using a deep thinking process within a single pass of a VLM. The synthetic data incorporates real-world images with accurate and relevant texts, ensuring fidelity through self-evaluation and refinement. The model's effectiveness underscores the potential of high-quality synthetic data in improving multimodal multilingual embeddings.

\smallskip
\noindent\textbf{VisRAG-Ret}~\cite{yu2025visrag} is a VLM-based retriever component of the VisRAG framework, designed to enhance retrieval-augmented generation by directly processing document images. Unlike traditional text-based RAG systems that rely on parsed text, VisRAG-Ret utilizes a VLM to embed document images, preserving the visual layout and content. It employs a bi-encoder architecture, mapping both the query and document images into a shared embedding space, facilitating efficient retrieval.



\section{Dataset Details}

In this section, we describe the datasets used in our experiments, covering both text-only and multimodal benchmarks. The text-only datasets are employed to evaluate retrieval and reasoning in purely linguistic settings, while the multimodal ones are used to assess cross-modal understanding and MRAG performance. Together, these datasets provide a comprehensive evaluation framework for analysing the effectiveness and generalization of our proposed method.

\smallskip
\noindent\textbf{HotPotQA}~\cite{yang2018hq} is a popular dataset for multi-hop question answering, comprising questions that require synthesizing information across multiple Wikipedia articles. The dataset includes complex query types, such as comparison and bridge questions. It contains 90{,}447 training samples, and we follow ARPO~\citep{dong2025arpo,dong2025aepo} use a held-out validation set with 250 examples for evaluation.

\smallskip
\noindent\textbf{2WikiMultihopQA}~\cite{ho20202wiki} is a large-scale dataset aimed at multi-hop reasoning, constructed by combining structured knowledge from Wikidata with unstructured passages from Wikipedia. It features diverse question formulations and annotated reasoning chains to facilitate explainable multi-step QA. The dataset includes 15{,}000 training samples, and our experiments use the test set consisting of 250 examples.

\smallskip
\noindent\textbf{Bamboogle}~\cite{press2023bamboogle} consists of manually curated multi-hop questions designed to test compositional reasoning. Some questions demand up to four inference steps, presenting a significant challenge in integrating information across multiple supporting facts. It provides only a test set, which we use for evaluation and which contains 125 examples.

\smallskip
\noindent\textbf{MuSiQue}~\cite{trivedi2022musique} focuses on sequential multi-hop inference, where each reasoning step depends on the output of the previous one. This dependency-based structure increases the difficulty of the task. The dataset comprises 19{,}938 training examples, and we use its development set with 250 held-out samples for evaluation.

\smallskip

For the four text-only datasets above, we construct two separate Wikipedia-derived corpora—one for training and one for evaluation. To build the training corpus, we aggregate all training questions and retrieve their top-20 relevant Wikipedia passages using the E5 retriever over the full Wikipedia dump. The retrieved passages are then deduplicated to form the final training corpus. Similarly, the evaluation corpus is constructed by collecting all test questions and retrieving their top-20 Wikipedia passages, followed by deduplication. The training corpus is used during the feedback collection stage, where \nyx-pretrained retrieves relevant passages to construct the ``downstream VLM preference dataset'' for fine-tuning. The evaluation corpus is used for benchmarking on the four text-only datasets during the final testing stage.

\smallskip
\noindent\textbf{MultimodalQA}~\cite{talmor2021mmqa} is a challenging question-answering dataset that necessitates joint reasoning across text, tables, and images. It includes 23{,}817 training examples and 2{,}411 testing examples. We combined text, tables, and images to create a large mixed-modal corpus containing 285{,}370 instances for this MMQA task.

\smallskip
\noindent\textbf{ScienceQA}~\cite{lu2022sciqa} is a large-scale multimodal science question dataset that annotates answers with detailed lectures and explanations. Each question is accompanied by context, either in the form of natural language or an image. For this dataset, we constructed two corpora: one consists of all the lectures appearing in the dataset, with duplicates removed to form the lecture corpus; the other contains the question-answer pairs from the training set, forming the example QA corpus. During testing, we retrieve one lecture and two example QAs to serve as external support information. The training set contains a total of 12{,}726 examples, while the testing set has 4{,}241 examples.

\section{Details for Raw QA Generation}

In this section, we provide additional details on the raw QA generation process described in Subsection~\ref{subsec:nyxqa}. Specifically, for a randomly sampled subset of 10{,}000 document instances from $\mathcal{C}_{\text{mix}}$, we employ two generation strategies depending on whether a document contains images. Using the InternVL3-78B model, we instruct it to produce up to five context-independent question–answer pairs for each document.

\bigskip
\noindent\textbf{Prompt for Text-Only Documents.}
The following prompt is used when the input document contains only textual content:

\begin{promptbox}[title=Instructions for text QA Pair Generation]
Given a text, please analyze the content of the text and raise no more than five questions along with their corresponding answers.\\
\textbf{Requirements:}\\
1.\quad The question must be independent of the context, that is, it cannot rely on background information that is not mentioned.\\
2.\quad The questions raised can be answered in concise language.

\textbf{Example:}

\textbf{Text:}
They broke the law, but it's not a felony. It's an act of love. It's an act of commitment to your family. I honestly think that that is a different kind of crime that there should be a price paid, but it shouldn't rile people up that people are actually coming to this country to provide for their families.
21 thoughts on ``Unethical Quote of the Month: Jeb Bush''

\textbf{Incorrect question:}
What does the speaker think about this crime? (Without specifying who the "speaker" is)

\textbf{Correct question:}
What type of crime does Jeb Bush describe as being committed by people coming to the country to provide for their families?

\textbf{Answer:}
Jeb Bush describes it as an act of love and commitment to family, not a felony.

\textbf{Output format:}
[Q1:... ,A1:... ], [Q2:... ,A2:... ], ...
\end{promptbox}

\vspace{1em}
\noindent\textbf{Prompt for Multimodal QA.}
When a document contains both text and images, the model is guided with the following prompt.

\begin{promptbox}[title=Instructions for multimodal QA Pair Generation]
You are given a document containing text and images, please analyze the content and raise no more than five questions along with their corresponding answers.  \\
\textbf{Requirements:}\\
1.\quad The question must be independent of the context, that is, it cannot rely on background information that is not mentioned.\\
2.\quad You can ask questions about the images in the document, but you need to clearly indicate them like:“Based on the image, \texttt{<image2>}, ...” or “Considering both images, \texttt{<image1>} and \texttt{<image3>}, ...” etc. \\
3.\quad The questions raised can be answered in concise language.

\textbf{Example:}

\textbf{Document:}
\texttt{<|image|>}The statement by Jeb Bush has its sunny side, I suppose: with any luck, it should ensure that we don’t have a Bush-Clinton contest in 2016. Maybe that was Jeb’s intent. Otherwise, his comments are irresponsible attacks on the rule of law, common sense, fairness and national sovereignty.

There are means by which we can control our border better than we have. And there should be penalties for breaking the law. But the way I look at this — and I’m going to say this, and it’ll be on tape and so be it. The way I look at this is someone who comes to our country because they couldn’t come legally, they come to our country because their families — the dad who loved their children — was worried that their children didn’t have food on the table. And they wanted to make sure their family was intact, and they crossed the border because they had no other means to work to be able to provide for their family. 

\textbf{Incorrect question:}
Considering both the text and \texttt{<image1>}, what might be the context of Jeb Bush's speech? (The question cannot be answered without context)

\textbf{Correct question:}
In the image, \texttt{<image1>}, what might be the context of Jeb Bush's speech? 

\textbf{Incorrect question:}
What is the main concern expressed about Jeb Bush's comments? (Without specifying what the "comments" is)

\textbf{Correct question:}
What is the main concern expressed about Jeb Bush's comments ``someone who comes to our country because they couldn't come legally, they come to our country because their families''?

\textbf{Output format:}
[Q1:... ,A1:... ], [Q2:... ,A2:... ], ...

\end{promptbox}